\documentclass[journal]{IEEEtran}
\ifCLASSINFOpdf
\else
\fi
\usepackage[T1]{fontenc}
\usepackage{bm}
\usepackage{times}
\usepackage{latexsym}
\usepackage{helvet}
\usepackage{couriers}
\usepackage{xcolor}
\usepackage[hyphens]{url}
\usepackage{graphicx}
\usepackage{multirow}
\usepackage{multicol}
\usepackage{array}
\usepackage{amsfonts}
\usepackage{algpseudocode}
\usepackage{booktabs}
\usepackage{amsmath}
\usepackage{cite}
\usepackage{multirow}
\usepackage{siunitx}
\usepackage{subcaption}
\usepackage[numbers]{natbib}

\begin{document}
%
\title{Dynamic Multi-Branch Layers for On-Device Neural Machine Translation}
%
%
%

\author{Zhixing Tan, Zeyuan Yang, Meng Zhang, Qun Liu, Maosong Sun, Yang Liu,~\IEEEmembership{Senior Member,~IEEE,}
\thanks{Z. Tan, Z. Yang, M. Sun, and Y. Liu are with the Department of Computer Science and Technology, Tsinghua University, Beijing 100084, China (e-mail: zxtan@tsinghua.edu.cn; yangzeyu21@mails.tsinghua.edu.cn; sms@tsinghua.edu.cn; liuyang2011@tsinghua.edu.cn). \textit{(Corresponding author: Yang Liu.)}}
\thanks{M. Zhang and Q. Liu are with Huawei Noah's Ark Lab. (e-mail: zhangmeng92@huawei.com; qun.liu@huawei.com).}
}

%
%

\markboth{IEEE/ACM Transactions on Audio, Speech, and Language Processing}%
{Shell \MakeLowercase{\textit{et al.}}: Bare Demo of IEEEtran.cls for IEEE Journals}
%



\maketitle

\begin{abstract}
With the rapid development of artificial intelligence (AI), there is a trend in moving AI applications, such as neural machine translation (NMT), from cloud to mobile devices. Constrained by limited hardware resources and battery, the performance of on-device NMT systems is far from satisfactory. Inspired by conditional computation, we propose to improve the performance of on-device NMT systems with dynamic multi-branch layers. Specifically, we design a layer-wise dynamic multi-branch network with only one branch activated during training and inference. As not all branches are activated during training, we propose shared-private reparameterization to ensure sufficient training for each branch. At almost the same computational cost, our method achieves improvements of up to 1.7 BLEU points on the WMT14 English-German translation task and 1.8 BLEU points on the WMT20 Chinese-English translation task over the Transformer model, respectively. Compared with a strong baseline that also uses multiple branches, the proposed method is up to 1.5 times faster with the same number of parameters.
\end{abstract}

\section{Introduction}

Machine translation is a classic problem of artificial intelligence (AI) that aims to translate natural languages automatically. With the rapid development of deep learning, neural machine translation (NMT)~\cite{sutskever2014sequence,bahdanau2015nmt,vaswani2017attention} has achieved great success and become the dominant approach to MT. Recently, there has been an increasing interest in moving AI applications, such as NMT, from cloud to mobile devices. Compared with cloud-based NMT services, on-device NMT systems offer increased privacy, low latency, and a more compelling user experience.

\begin{figure}[!t]
\centering
\includegraphics[width=0.45\textwidth]{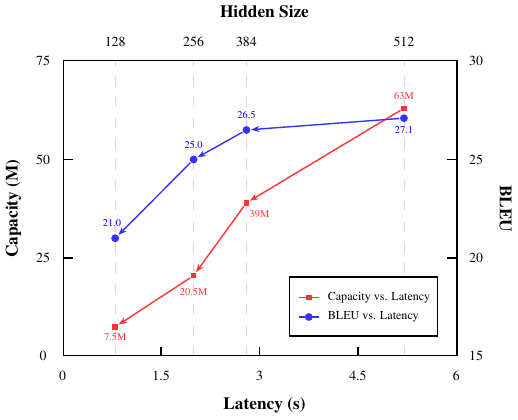}
\caption{Latency, performance (BLEU), and capacity of Transformer models with different hidden sizes on a Raspberry Pi 4 device. All models are trained on the WMT14 English-German dataset. Scaling down the hidden size of the model (from 512 to 128) reduces the translation latency, but also lowers the model's capacity and hurts the translation performance.}\label{fig:latency}
\end{figure}

However, it is challenging to deploy NMT models on edge devices. Due to high computation cost, on-device NMT systems face a trade-off between latency and performance~\cite{wu2020lite}. Figure~\ref{fig:latency} gives an illustration of the latency, performance, and capacity of Transformer models with different hidden sizes when translating a sequence with 30 tokens on a Raspberry Pi 4 device. As we can see, a Transformer-base model takes over 5 seconds to translate a sequence of 30 tokens and such long latency is not desired for real-time applications. Although the latency can be reduced by simply scaling down the hidden size of the network, it also weakens the model's capacity, making the translation performance of on-device NMT models far from satisfactory. As the model capacity is a key factor in determining the performance of neural networks~\cite{shazeer2017outrageously}, how to increase the capacity of on-device NMT models without sacrificing efficiency is an important problem for achieving a better trade-off between latency and performance.

This problem has received increasing attention in the community in recent years \cite{bengio2013estimating,shazeer2017outrageously,bapna2020controlling,wu2020lite,chung2020extremely}. Among prior studies, conditional computation~\cite{bengio2013estimating,cho2014exponentially,bengio2015conditional}, which proposes to activate parts of a neural network in an input-dependent fashion, is a representative method for improving model capacity without a proportional increase in computation time. Recently, \citet{shazeer2017outrageously} have proposed a sparsely-gated mixture-of-experts (MoE) approach. While their approach dramatically increases model capacity by introducing multiple alternatives, it still inevitably incurs non-negligible computational overhead. \citet{bapna2020controlling} propose conditional computation Transformer, which can dynamically skip a sub-layer of the model based on the complexity of input and demonstrates its effectiveness over the vanilla Transformer within a small computational budget on an average basis. However, the computational time may vary significantly when translating sequences with the same length as in the worst case no sub-layer can be skipped, which is not desired for on-device NMT services. As a result, how to use conditional computation to improve model capacity without sacrificing efficiency still remains a major challenge for on-device NMT.

\begin{figure}[!t]
\centering 
    \begin{subfigure}[b]{0.23\textwidth}
        \centering
        \includegraphics[width=\textwidth]{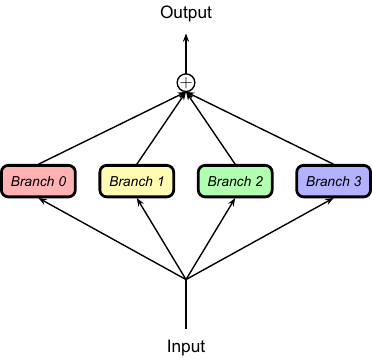}
        \caption{}
    \end{subfigure}
    \hfill
    \begin{subfigure}[b]{0.23\textwidth}
        \centering
        \includegraphics[width=\textwidth]{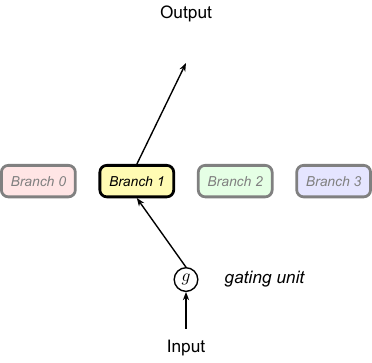}
        \caption{}
    \end{subfigure}
    \caption{(a) A conventional multi-branch layer and (b) a dynamic multi-branch layer. By dynamically activating only one branch during training and inference via an input-dependent gate, our goal is to enable on-device NMT to share the same model capacity with conventional multi-branch networks and maintain the same efficiency with standard single-branch networks.}\label{fig:mb}
\end{figure}

Along the line of conditional computation, we propose to use {\em dynamic multi-branch} (DMB) layers for on-device NMT. As shown in Figure~\ref{fig:mb}, a dynamic multi-branch layer is capable of dynamically activating a single branch using an input-sensitive gate, enabling the resulting NMT model to have increased capacity thanks to the use of more branches while keeping the same efficiency with the standard single-branch model. To ensure sufficient training of each branch, we propose shared-private reparameterization that requires zero computational and memory overhead during inference. Based on DMB layers, we also propose Transformer-DMB architecture, which extends both the feed-forward and attention layers of a Transformer model with DMB counterparts, to effectively increase the capacity of the model while keeping nearly the same efficiency.

We conduct experiments on the WMT14 English-German (En-De) and the WMT20 Chinese-English (Zh-En) translation tasks to verify the effectiveness of our proposed method. Experiments show that our method achieves improvements of up to 1.7 BLEU points on the WMT14 En-De translation task and up to 1.8 BLEU points on the WMT20 Zh-En translation task over the Transformer model while keeping nearly the same computation efficiency. The proposed method is also up to 1.5 times faster than a strong MoE-based baseline with the same number of parameters and comparable translation quality.

In summary, our contributions are:
\begin{itemize}
\item We propose \emph{dynamic multi-branch layers}, which is a simple yet effective method to increase the capacity of a network without sacrificing speed.
\item We propose \emph{shared-private reparameterization} to ensure the effective training of each branch in a DMB layer.
\item We propose \emph{Transformer-DMB architecture} that significantly increases the capacity while keeping nearly the same computation cost of the Transformer model.
\end{itemize}

\section{Background}
\subsection{The Transformer Model}
Transformer~\cite{vaswani2017attention} is the state-of-the-art neural architecture for machine translation. The encoder of the Transformer typically consists of 6 layers, where each encoder layer is composed of a multi-head attention (MHA) sub-layer and a feed-forward (FFN) sub-layer. The decoder of the Transformer also consists of 6 layers, where each decoder layer is composed of a masked MHA sub-layer, an encoder-decoder MHA sub-layer, and an FFN sub-layer.

Multi-head attention consists of $H$ parallel heads, each of which is a scaled dot-product attention. Given a set of $n$ queries $\mathbf{Q} \in \mathbb{R}^{n \times d}$, $m$ keys $\mathbf{K}  \in \mathbb{R}^{m \times d}$, and $m$ values $\mathbf{V} \in \mathbb{R}^{m \times d}$, where $d$ is the hidden size, the output of an MHA sub-layer is a combination of the outputs of each head. Formally, it can be described as
\begin{align}
  \mathrm{MHA}(\mathbf{Q}, \mathbf{K}, \mathbf{V}) &= \mathrm{Concat} (\mathrm{head}_1, ..., \mathrm{head}_H) \mathbf{W}_o,
\end{align}
where $\mathbf{W}_o \in \mathbb{R}^{d \times d}$ are learnable parameters. Each attention head attends to different representation subspaces at different positions:
\begin{align}
  \mathrm{head}_h &= \mathrm{Attention}\large(\mathbf{Q}\mathbf{W}_{q}^{(h)}, \mathbf{K}\mathbf{W}_{k}^{(h)}, \mathbf{V}\mathbf{W}_{v}^{(h)}\large),
\end{align}
where $\mathbf{W}_{q}^{(h)} \in \mathbb{R}^{d \times d_h}, \mathbf{W}_{k}^{(h)} \in \mathbb{R}^{d \times d_h}$, $\mathbf{W}_{v}^{(h)} \in \mathbb{R}^{d \times d_h}$, and $d_h = d / H$. The computation involved in scaled dot-product attention network can be described as
\begin{align}
    \mathrm{Attention}(\mathbf{Q}, \mathbf{K}, \mathbf{V}) &= \mathrm{softmax} \left( \frac{\mathbf{Q} \mathbf{K}^{\top}}{\sqrt{d_k}} \right) \mathbf{V}.
\end{align}
To compensate the attention's negligence of the order of input sequence, the Transformer adds \emph{positional encoding} to the bottom of both the encoder and the decoder. Please refer to \cite{vaswani2017attention} for more details.

The FFN sub-layer consists of two linear layers with ReLU as the activation function. Formally, given an input vector $\mathbf{x} \in \mathbb{R}^d$, the output of an FFN sub-layer can be described as
\begin{align}
  \mathrm{FFN}(\mathbf{x}) &= \mathbf{W}_2 \mathrm{ReLU}(\mathbf{W}_1\mathbf{x}),
\end{align}
where $\mathbf{W}_1 \in \mathbb{R}^{d_f \times d}, \mathbf{W}_2 \in \mathbb{R}^{d \times d_f}$, and $d_f$ is the hidden filter size.

The loss function for learning the Transformer model is the cross-entropy loss. Given a source sequence $\mathbf{x}$ of length $S$ and a target sequence $\mathbf{y}$ of length $T$, the loss function $\mathcal{L}_{m}$ is defined as
\begin{align}
\mathcal{L}_{m} = -\frac{1}{T}\sum_{t=1}^{T}\log P(\mathbf{y}_t|\mathbf{x}). \label{eq:ce}
\end{align}

\subsection{Sparsely-Gated Mixture-of-Experts Layer}
Sparsely-gated mixture-of-experts (MoE) layer~\cite{shazeer2017outrageously} is a method to realize the promise of conditional computation. Each MoE layer consists of a set of $N$ experts $f_i$ and a gating network $g$. Given an input vector $\mathbf{x} \in \mathbb{R}^d$, the output $\mathbf{y}$ of the MoE layer is formally described as follows:
\begin{align}
\mathbf{y} = \sum_{i=1}^N g(\mathbf{x}, i) \cdot f_i(\mathbf{x}),
\end{align}
where ``$\cdot$'' denotes the scalar-matrix multiplication, and $g(\mathbf{x}, i)$ denotes the $i$-th element in $g(\mathbf{x})$.

To increase the number of experts $N$ without a proportional increase of computation, \citet{shazeer2017outrageously} introduced a noisy top-$k$ gating mechanism. Formally, the computation involved in $g(\cdot)$ can be described as
\begin{align}
g(\mathbf{x}) &= \mathrm{softmax}(\mathrm{KeepTopK}(H(\mathbf{x}), k)),\label{eq:moe_g} \\
H(\mathbf{x}) &= \mathbf{W}\mathbf{x} + \bm{\epsilon} \odot \mathrm{softplus}(\mathbf{W}_{n}\mathbf{x)},
\end{align}
where $\bm{\epsilon} \sim \mathcal{N}(\mathbf{0}, \mathbf{1})$ is a $d$-dimensional noise vector and ``$\odot$'' denotes the element-wise product. ``$\mathrm{softplus}$'' is an activation function which is formally defined as
\begin{align}
\mathrm{softplus}(x) = \log (1+\exp(x)).
\end{align}
The $i$-th element in the vector $\mathrm{KeepTopK}(\mathbf{v}, k)$ is formally defined as
\begin{align}
\mathrm{KeepTopK}(\mathbf{v}, k, i) &= \left\{
    \begin{array}{ll}
      \mathbf{v}_i & \hspace{0.3em}\textrm{if $\mathbf{v}_i$ is in the top $k$}\\
          & \hspace{0.3em}\textrm{elements in $\mathbf{v}$,} \\
      -\infty & \hspace{0.3em}\textrm{otherwise.}
    \end{array}\right.
\end{align}

When setting $k>1$, gradients can back-propagate through the noise top-$k$ gating network to its inputs. As a result, the MoE layers can be learned in an end-to-end manner. \emph{However, gradients can no longer back-propagate through the network to its inputs when $k=1$}. The reason is that $g(\mathbf{x})$ is always a constant vector regardless of the input $\mathbf{x}$ in Eq~(\ref{eq:moe_g}). Therefore, to enable end-to-end training, the MoE layers still introduce non-negligible computational overhead.

\citet{shazeer2017outrageously} also introduced two additional losses to each MoE layer. The first loss is the diversity loss, which encourages all experts to have equal importance. The second loss is the load balance loss, which ensures balanced loads when training on distributed hardware.

\section{Proposed Method}
In this section, we first describe the definition of dynamic multi-branch layers in Section~\ref{sec:dmb}. Then we describe the shared-private reparameterization in Section~\ref{sec:spr}. Finally, we describe the architecture of our models in Section~\ref{sec:arch}.

\subsection{Dynamic Multi-Branch Layers}~\label{sec:dmb}
\begin{figure}[!t]{}
\centering
\includegraphics[width=0.25\textwidth]{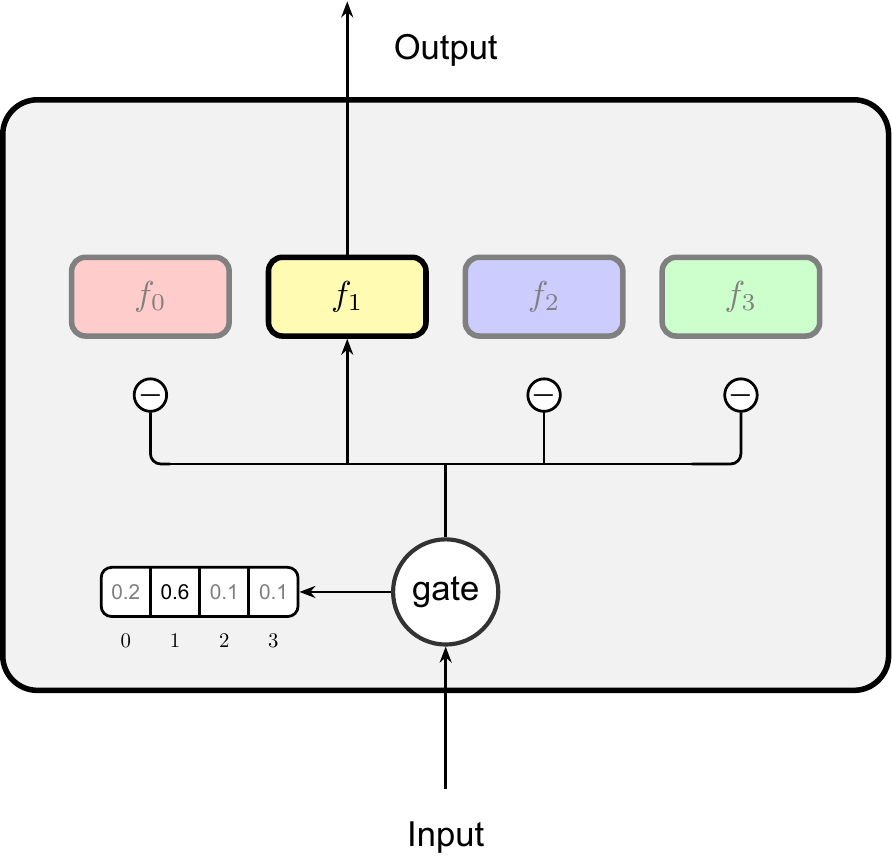}
\caption{An overview of a dynamic multi-branch layer with 4 branches. Based on the predictions of the gating unit, only one branch in a layer is active during training and inference.}\label{fig:dmb}
\end{figure}

We propose \emph{dynamic multi-branch} (DMB) layers for increasing the model capacity while maintaining roughly the same computational cost. Similar to MoE layers, each DMB layer consists of a set of $N$ identical branches with different parameters. We use $d$ to denote the hidden size of the network. Given an input vector $\mathbf{x} \in \mathbb{R}^d$, we denote the $i$-th branch in the layer as $f_i(\mathbf{x};\bm{\theta}_i)$, where $f_i$ can be an arbitrary differentiable function with parameters $\bm{\theta}_i$. The output of a DMB layer is described as

\begin{align}
\mathbf{y} &= \sum_{i=1}^{N} g(\mathbf{x}, i) \cdot f_i(\mathbf{x};\bm{\theta}_i).
\end{align}

Different from MoE layers where $g(\mathbf{x}, i)$ is a real-valued scalar, $g(\mathbf{x}, i)$ in DMB layers is a \emph{binary-valued} scalar indicating whether the $i$-th branch is activated or not. The value of $g(\mathbf{x}, i)$ is dynamically computed based on the input $\mathbf{x}$. For each DMB layer, we first employ a lightweight gating unit to learn a probability distribution for activating each branch. We use $a(\mathbf{x})$ to denote the predicted distribution and $a_i(\mathbf{x})$ to denote the $i$-th element in $a(\mathbf{x})$. The gating unit is simply a linear layer followed by a softmax activation function:
\begin{align}
a(\mathbf{x}) = \mathrm{softmax}(\mathbf{W}_g\mathbf{x} + \mathbf{b}_g),
\end{align}
where $\mathbf{W}_g \in \mathbb{R}^{N \times d}$ and $\mathbf{b} \in \mathbb{R}^N$ are learnable parameters. Then we activate the branch with the highest probability. Formally, the value of $g(\mathbf{x}, i)$ is described as
\begin{equation}
  g(\mathbf{x}, i) = \left\{
    \begin{array}{ll}
      1 & \quad\textrm{if $a_i(\mathbf{x})$ is the top}\\
          & \quad\textrm{element in $a(\mathbf{x})$,} \\\\
      0 & \quad\textrm{otherwise.}
    \end{array}\right.
\end{equation}

Figure~\ref{fig:dmb} gives an illustration of a DMB layer. By introducing the gating unit, we can ensure that only one branch in a layer is active during training and inference. Unfortunately, binary outputs of the gating units are not differentiable, which complicates the learning of DMB layers. To enable the end-to-end training of gating units, we introduce two auxiliary losses: the diversity loss and the entropy loss.

The first auxiliary loss is the \emph{diversity loss}. The goal of diversity loss is to encourage a balanced utilization of each branch. Following \citet{shazeer2017outrageously}, we minimize the coefficient of variation among all branches. For a batch of $M$ inputs $\{ \mathbf{x}_i \}$, the loss function is formally described as
\begin{align}
  \mu &= \frac{1}{N} \sum_{i=1}^{N}\sum_{j=1}^{M} a_i(\mathbf{x}_j), \\
  \sigma^2 &= \sum_{i=1}^{N} (\sum_{j=1}^{M} a_i(\mathbf{x}_j) - \mu)^2, \\
  \mathcal{L}_{d} &= \frac{\sigma^2}{\mu^2}.
\end{align}

The second auxiliary loss is the \emph{entropy loss}. We expect that the gating unit can give a high probability when activating a branch. Therefore, we minimize the entropy of predicted distribution $a(\mathbf{x})$. Formally, entropy loss is defined as
\begin{align}
\mathcal{L}_{e} = -\frac{1}{M}\sum_{j=1}^{M}\sum_{i=1}^{N} a_i(\mathbf{x}_j) \cdot \log a_i(\mathbf{x}_j).
\end{align}

By using the auxiliary losses, the gating units can receive gradients directly from the loss functions. Despite of its simplicity, in our experiments, we found the two auxiliary losses are sufficient for learning the gating units.

Therefore, DMB layers achieve nearly the same computational efficiency as the corresponding single-branch layers. The only additional computational overhead of a DMB layer is calculating $a(\mathbf{x})$, which is negligible compared with the computation of an active branch $f_i$.

\subsection{Shared-Private Reparameterization}~\label{sec:spr}
Ideally, we expect a balanced utilization for all branches of a DMB layer. However, in this situation, each branch is only trained with a subset of training examples. We refer to this phenomenon as the \emph{shrinking training examples} problem. For example, for a smaller training set or large choices of $N$, each branch is only trained with $1/N$ examples in expectation, which often leads to insufficient training of each branch.

To alleviate this problem, we propose a method called \emph{shared-private reparameterization}: for a given DMB layer, the parameters $\bm{\theta}_i$ of the $i$-th branch $f_i$ is composed of two separate set of parameters $\{\bm{\theta}_S, \bm{\theta}_{P_i}\}$. $\bm{\theta}_S$ is the \emph{shared parameters} for all branches in the DMB layer. In contrast, $\bm{\theta}_{P_i}$ is the \emph{private parameters} bound to the $i$-th branch in the DMB layer. $\bm{\theta}_i$ is reparameterized as a summation of $\bm{\theta}_S$ and $\bm{\theta}_{P_i}$:
\begin{align}\label{eq:sp}
\bm{\theta}_i = \bm{\theta}_S + \bm{\theta}_{P_i}.
\end{align}

Suppose the gating unit chooses the $j$-th branch in the forward pass. During the backward-pass, the gradients with respect to $\bm{\theta}_S$ and $\bm{\theta}_{P_i}$ are
\begin{align}
\frac{\partial \mathcal{L}}{\partial \bm{\theta}_S} &= \frac{\partial \mathcal{L}}{\partial \mathbf{y}}  \frac{\partial f_j(\mathbf{x};\bm{\theta}_j)}{\partial \bm{\theta}_j}, \\
\frac{\partial \mathcal{L}}{\partial \bm{\theta}_{P_i}} &= \frac{\partial \mathcal{L}}{\partial \mathbf{y}}  \frac{\partial f_j(\mathbf{x};\bm{\theta}_j)}{\partial \bm{\theta}_{j}} \frac{\partial \bm{\theta}_{j}}{\partial \bm{\theta}_{P_i}}.
\end{align}
The shared parameters are always updated no matter which branch is chosen. In contrast, the private parameters $\bm{\theta}_{P_i}$ receive non-zero gradients only when $i$ is equal to $j$. By introducing the shared-private reparameterization, we expect that each branch in a DMB layer is not only able to learn shared knowledge with $\bm{\theta}_S$, but also preserves distinct characteristics via $\bm{\theta}_{P_i}$.

Another benefit of using a summation in Equation~(\ref{eq:sp}) is that we can easily \emph{pre-compute} $\bm{\theta}_i$ when the training is over. As a result, there is zero computational and memory overhead after training when introducing $\bm{\theta}_S$.

\begin{figure}[!t]{}
\centering
\includegraphics[width=0.4\textwidth]{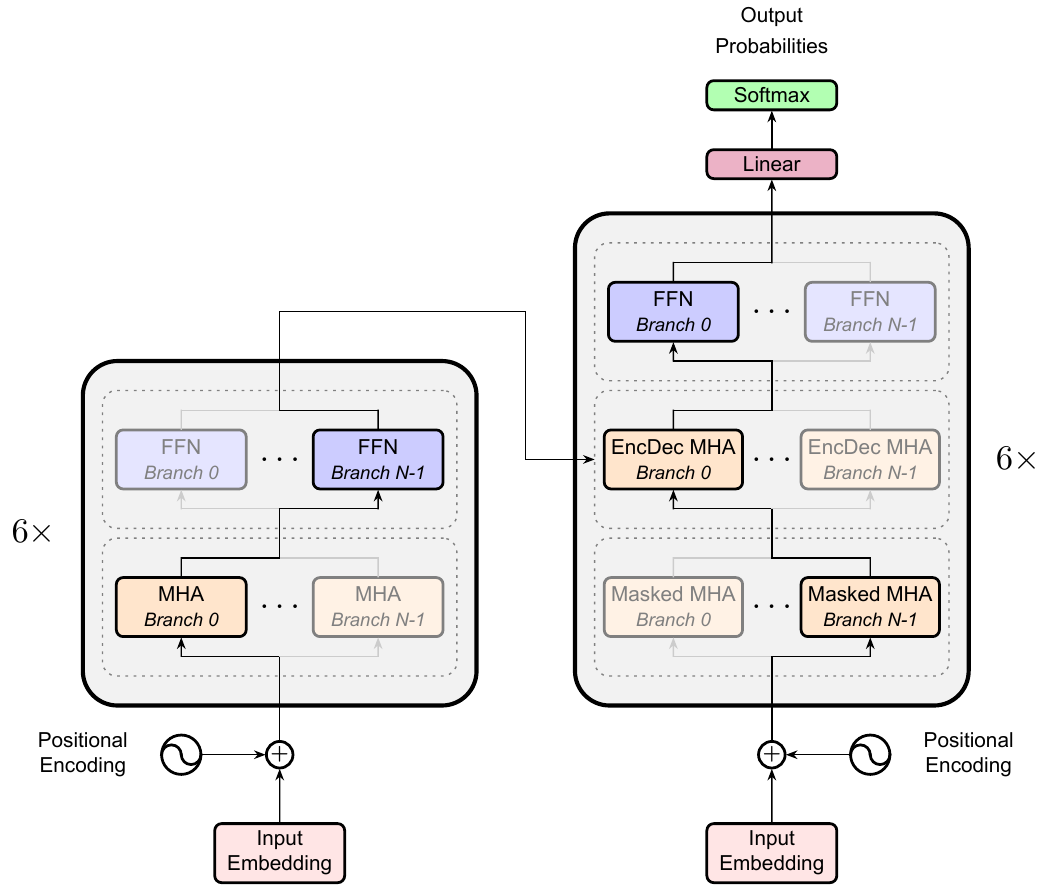}
\caption{Transformer with dynamic multi-branch layers.}\label{fig:arch}
\end{figure}

\subsection{Architecture and Training Objective}~\label{sec:arch}
We extend the Transformer architecture with DMB layers, which we refer to as the Transformer-DMB architecture. Figure~\ref{fig:arch} depicts the Transformer-DMB architecture. Different from ~\cite{shazeer2017outrageously} where MoE layers are only applied for FFN layers, we apply DMB at a more fine-grained level. For DMB FFN layers, we introduce a single gating unit $g_f$. Suppose the gate predictions $g_f(\mathbf{x}, k) = 1$ for a given input $\mathbf{x}$, the computation of DMB FFN layers is formally defined as:
\begin{align}
\mathrm{FFN}_{\textrm{DMB}}(\mathbf{x}) = \mathbf{W}_2^{(k)} \mathrm{ReLU}(\mathbf{W}_1^{(k)}\mathbf{x}),
\end{align}
where $\mathbf{W}_1^{(k)}, \mathbf{W}_2^{(k)}$ are parameters of the $k$-the branch of the DMB FFN layer.

For DMB MHA layers, we also introduce an additional gating unit $g_a$. Given a set of $n$ inputs $\mathbf{X} = [\mathbf{x}_1, \ldots, \mathbf{x}_n]$ and a set of $N$ weights $\{ \mathbf{W}^{(k)} \}$, we use $g(\mathbf{X}, \{\mathbf{W}^{(k)}\})$ to denote the following computations:
\begin{align}
  g(\mathbf{X}, \{\mathbf{W}^{(k)}\}) = [\mathbf{W}^{(i_1)}\mathbf{x}_1, \ldots, \mathbf{W}^{(i_n)}\mathbf{x}_n],
\end{align}
where $g(\mathbf{x}_k, i_k) = 1$ for $k \in \{1, \ldots, n\}$ and $i_k \in \{1, \ldots, N\}$. Then the computations involved in DMB MHA layers can be formally described as
\begin{align}
  \mathrm{MHA}_{\textrm{DMB}}(\mathbf{Q}, \mathbf{K}, \mathbf{V}) = g_a(\mathbf{H}, \{\mathbf{W}_o^{(k)}\}),
\end{align}
where $\mathbf{W}_o^{(k)}$ are weights associated with the $k$-th MHA branch, $\mathbf{H}=\mathrm{Concat} (\mathrm{head}_1, ..., \mathrm{head}_H)$, and
\begin{align}
  \mathrm{head}_h = \mathrm{Attention}\large(\mathbf{Q}^{(h)}, \mathbf{K}^{(h)}, \mathbf{V}^{(h)}\large).
\end{align}
Furthermore, we have
\begin{align}
  \mathbf{Q}^{(h)} &= g_a(\mathbf{Q}, \{\mathbf{W}_q^{(h,k)}\}), \\
  \mathbf{K}^{(h)} &= g_a(\mathbf{K}, \{\mathbf{W}_k^{(h,k)}\}), \\
  \mathbf{V}^{(h)} & = g_a(\mathbf{V}, \{\mathbf{W}_v^{(h,k)}\}),
\end{align}
where $\mathbf{W}_q^{(h,k)}$, $\mathbf{W}_k^{(h,k)}$, and $\mathbf{W}_v^{(h,k)}$ are weights associated with the $h$-th head of the $k$-th MHA branch.

Suppose there are a total of $L$ DMB layers. For the $i$-th gating unit $g(\cdot, i)$, the associated auxiliary losses $\mathcal{L}_{d,i}$ and $\mathcal{L}_{e,i}$ are added to the original cross-entropy loss $\mathcal{L}_{m}$ (see Eq.~(\ref{eq:ce})) with a weight $\alpha$. As a result, the final loss function of our model becomes
\begin{equation}
\mathcal{L} = \mathcal{L}_{m} + \alpha \frac{1}{L}\sum_{i=1}^{L}(\mathcal{L}_{d,i} + \mathcal{L}_{e,i}).
\end{equation}

\begin{table*}[t]
\centering
\begin{tabular}{l rrrr rrrr}
\toprule
\multirow{2}{*}{\bf Method} & \multicolumn{4}{c}{\bf En-De} & \multicolumn{4}{c}{\bf Zh-En} \\
\cmidrule(lr){2-5} \cmidrule(lr){6-9}
& \textit{\#Params.} & \textit{\#MA.} & \textit{BLEU} & \textit{PTR} & \textit{\#Params.} & \textit{\#MA.} & \textit{BLEU} & \textit{PTR} \\
\midrule
Transformer (tiny)~\tiny{\cite{vaswani2017attention}} & 7.5M & 229.0M & 21.0 & 13.9 & 11.0M & 209.7M & 19.0 & 13.2 \\
Evolved Transformer~\tiny{\cite{so2019evolved}} & 15.9M & - & 22.0 & - & - & - & - & - \\
Lite Transformer~\tiny{\cite{wu2020lite}} & 16.0M & 286.6M & 22.6 & 13.3 & - &- & - & - \\
Transformer-MoE (tiny) & 15.8M & 313.2M & 22.5 & 12.7 & 19.3M & 293.9M & 21.0 & 12.2 \\
Transformer-DMB (tiny) & 15.8M & 229.6M & 22.7 & \textbf{15.0} & 19.3M & 210.3M & 20.8 & \textbf{14.3} \\ 
\midrule
Transformer (small)~\tiny{\cite{vaswani2017attention}} & 20.5M & 623.2M & 25.0 & 10.0 & 27.4M & 584.6M & 24.3 & 10.1 \\
Lite Transformer~\tiny{\cite{wu2020lite}} & 38.5M & 761.1M & 25.6 & 9.3 & - & - & - & - \\
HAT\tiny{~\cite{wang2020hat}} & 80.0M & 2.1G & 25.9 & 5.7 & - & - & - & - \\
Transformer-MoE (small) & 53.7M & 956.6M & 25.7 & 8.3 & 60.6M & 918.1M & 25.0 & 8.3 \\
Transformer-DMB (small) & 53.7M & 624.3M & 25.7 & \textbf{10.3} & 60.6M & 585.7M & 24.8 & ~~\textbf{10.2} \\
\bottomrule
\end{tabular}
\caption{Experiments on WMT14 En-De and WMT20 Zh-En datasets. ``\#Params.'' denotes the number of parameters, ``\#MA.'' denotes the number of Multi-Adds, and ``PTR'' denotes the performance-time ratio. Different from \citet{wu2020lite} and \citet{wang2020hat}, we also count the number of parameters and Mult-Adds in the embedding and classification layers. ``-'' indicates that the comparison is not applicable. The number of experts/branches $N$ is set to 4 for MoE and DMB.}
\label{tab:result}
\end{table*}

\section{Experiments}

\subsection{Setup}
\subsubsection{Datasets}
We evaluate our proposed models on English-German (En-De) and Chinese-English (Zh-En) translation tasks. The evaluation metric is case-sensitive BLEU~\cite{papineni2002bleu}. We use the number of Mult-Adds~\footnote{We use the \texttt{torchprofile} library to count the number of Mult-Adds.} to measure the computational costs of an NMT model~\cite{wu2020lite}. To quantify how well an on-device NMT system increases capacity while keeping efficiency, we introduce a measure called {\em performance-time ratio} (PTR) defined as
\begin{align}
\mathrm{PTR} = \frac{\textrm{BLEU}}{\sqrt{\textrm{\#Mult-Adds}}} \times 10^4.
\end{align}

For the En-De translation task, we use the WMT14 training corpus which contains 4.5M sentence pairs with 103M English words and 96M German words. We use a shared source-target vocabulary of about 37,000 tokens encoded by BPE~\cite{sennrich2016bpe}. We use \texttt{newstest2013} as the validation set and use \texttt{newstest2014} as the test set. We report the tokenized BLEU score as calculated by \texttt{multi-bleu.perl} to be in accordance with previous works.

For the Zh-En translation task, we use the training corpus provided by WMT20. After filtering duplicate entries, the corpus consists of 21M sentence pairs with 417M Chinese words and 452M English words. We use 32K BPE operations to build vocabularies. The validation set is \texttt{newsdev2017} and the test set is \texttt{newstest2020}. We report the detokenized BLEU score calculated by \textsc{SacreBLEU}~\footnote{Signature: BLEU+case.mixed+numrefs.1+smooth.exp+tok.13a\\+version.1.4.14}~\cite{post2018call}.

\subsubsection{Settings}

We compare our proposed models with the following baselines:
\begin{itemize}
\item \textit{Transformer}~\cite{vaswani2017attention}. State-of-the-art self-attention based neural machine translation architecture.
\item \textit{Transformer-MoE}. We implement the sparsely-gated MoE layer~\cite{shazeer2017outrageously} on top of the Transformer architecture. The overall architecture is similar to ours, with two key differences. First, we mix top-2 experts from a total of $N$ in the MoE architecture for end-to-end training of the gating unit. Second, we do not use the shared-private reparameterization in this approach.
\end{itemize}

Besides these baselines, we also compare our model with recent works on efficient NMT models such as Lite Transformer~\cite{wu2020lite} whenever possible.

We experiment with two commonly used settings for on-device NMT:
\begin{itemize}
\item \textit{Tiny setting}. We set the hidden size $d$ of the model to 128. The filter size of feed-forward layers is set to 512. Under this setting, the Transformer has 229.0M Mult-Adds for the En-De translation task.
\item \textit{Small setting}. A larger setting for more capable devices, which resulting around 623.2M Mult-Adds for a Transformer model trained on the En-De translation task. We set the hidden size $d$ of the model to 256. The filter size of feed-forward layers is set to 1,024.
\end{itemize}

For all our models, we use Adam~\cite{kingma2014adam} ($\beta_1$ = 0.9, $\beta_2$ = 0.98 and $\epsilon$ = 1$\times$ $10^{-9}$) as the optimizer. We empirically set $\alpha=0.1$ for our MoE and DMB approaches, and use $N=4$ experts or branches. For DMB layers, we initialize all shared parameters to zero at the beginning of training. Each mini-batch contains approximately 64K source and 64K target tokens. All other settings are the same with ~\citet{vaswani2017attention}. We train all models on a machine with 8 RTX 2080Ti GPUs for 10,000 steps and save a checkpoint every 1,000 steps. For all models, we averaged the last five checkpoints to obtain a single model. Decoding is performed using beam search. Following \cite{vaswani2017attention}, we set the beam size to 4 in all our experiments. We set the length penalty to 0.6 for En-De translation and 1.0 for Zh-En translation.

\subsection{Results}

\begin{table}[!t]
\centering
\begin{tabular}{lcc}
\toprule
\textbf{Model} & \textbf{Latency} & \textbf{BLEU} \\\midrule
\multicolumn{3}{c}{\textit{greedy search}}  \\\midrule
Transformer (tiny)~\tiny{\cite{vaswani2017attention}} & 0.39s & 20.0 \\
Transformer-MoE (tiny) & 0.60s & 21.5 \\
Transformer-DMB (tiny) & 0.42s & 21.8 \\
\midrule
Transformer (small)~\tiny{\cite{vaswani2017attention}} & 1.37s & 24.0 \\
Transformer-MoE (small) & 2.26s & 24.6 \\
Transformer-DMB (small) & 1.52s & 24.6 \\\midrule
\multicolumn{3}{c}{\textit{beam search}}  \\\midrule
Transformer (tiny)~\tiny{\cite{vaswani2017attention}} & 0.82s & 21.0 \\
Transformer-MoE (tiny) & 1.33s & 22.5 \\
Transformer-DMB (tiny) & 0.99s & 22.7 \\
\midrule
Transformer (small)~\tiny{\cite{vaswani2017attention}} & 1.95s & 25.0 \\
Transformer-MoE (small) & 4.15s & 25.7 \\
Transformer-DMB (small) & 2.72s & 25.7 \\
\midrule
\end{tabular}
\caption{Raspberry Pi 4 ARM CPU latency and BLEU comparisons with different models on the WMT14 En-De translation task. The number of experts/branches $N$ is set to 4 for MoE and DMB.}\label{tab:latency}
\end{table}

Table~\ref{tab:result} shows the results on the En-De and the Zh-En translation tasks, respectively. 

On the En-De translation task, our model outperforms the Transformer model by 1.7 BLEU points under tiny model settings, and 0.7 BLEU points under small model settings. The results of the Zh-En translation task are similar. Our model outperforms the Transformer model over 1.8 BLEU points under tiny model settings, and 1.0 BLEU point under small model settings. The performance of our model is on par with the Transformer model extended with MoE. The Transformer-DMB and Transformer-MoE models also have the same number of parameters. However, the computational overhead of our model is much less than the MoE approach. Our approach only adds 0.6M computational overhead for the tiny model compared to the corresponding Transformer architecture whereas the MoE approach adds 84.2M Mult-Adds. Under the small model settings, our Transformer-DMB model is 1.5 times faster than the Transformer-MoE model (624.3M Mult-Adds vs. 956.6M Mult-Adds), which is, therefore, more preferable for edge devices. We also compare our Transformer-DMB models with other models, such as Evolved Transformer~\cite{so2019evolved}, Lite Transformer~\cite{wu2020lite}, and HAT~\cite{wang2020hat}. With comparable performance, our models achieve the best performance-time ratio. Nevertheless, we believe our work is also applicable to Lite Transformer and HAT.

We report the inference latency of a sequence with 30 tokens in Table~\ref{tab:latency} on a Raspberry 4 device, which has 4 Cortex-A72 cores at 1.5GHz and 8GB RAM. For greedy search, the Transformer-DMB model costs about 7.7\% more time than the Transformer model under the tiny model settings and costs about 10.9\% more time under the small model settings. Compared with Transformer-MoE models, the Transformer-DMB model is about 1.4 times faster under the tiny model settings, and 1.5 times faster under the small model settings. For beam search, Transformer-DMB modes introduce more computational burdens to Transformer models compared with the results when using greedy search. This is because our current implementation splits a batched matrix into smaller matrices to enable conditional computations, which reduces the degree of parallelism for matrix multiplications. We believe the gap can be significantly narrowed with a dedicated DMB-aware matrix multiplication kernel.

\subsection{Ablation Study}
We conduct an ablation study on the WMT14 En-De translation task to investigate the performance of each component of our model, the results are reported in Table~\ref{tab:abl}. Based on the results, we have the following observations:
\begin{itemize}
\item Replacing all DMB MHA layers in the Transformer-DMB model with plain MHA layers lowers the BLEU score by 0.7 points. The results suggest that besides extending FFN layers with DMB FFN layers, extending MHA layers on both the encoder and the decoder with DMB MHA layers can further increase the capacity of the model, and is helpful for improving the translation performance.
\item Without using shared-private reparameterization, the Transformer-DMB model achieves 21.9 BLEU points, which underperforms the result of a model trained with shared-private reparameterization by 0.8 points. This shows that shared-private reparameterization is helpful in improving the learning of each branch in DMB layers.
\item Without introducing auxiliary losses (diversity loss or entropy loss or both) during training, the Transformer-DMB model only achieves around 21.1 BLEU points, which is nearly the same as a Transformer model with the same hidden size (21.0 BLEU points). The results indicate that both diversity and entropy loss are essential for learning DMB layers.
\end{itemize}

\begin{table}[!t]
\centering
\begin{tabular}{lc}
\toprule
\textbf{Method} & \textbf{BLEU} \\\midrule
Transformer-DMB (tiny) & 22.7 \\
\quad $-$ MHA DMB Layers & 22.0 \\
\quad $-$ Shared-Private Reparameterization & 21.9 \\
\quad\quad $-$ Diversity Loss & 21.0 \\
\quad\quad $-$ Entropy Loss & 21.1 \\
\quad\quad $-$ Auxiliary Losses & 21.1 \\
\bottomrule
\end{tabular}
\caption{Ablation study of our proposed Transformer-DMB model on the WMT14 En-De translation task.}\label{tab:abl}
\end{table}

\subsection{Effect of Reparameterzation and the Number of Branches}

\begin{figure}[!t]{}
\centering
\includegraphics[width=0.4\textwidth]{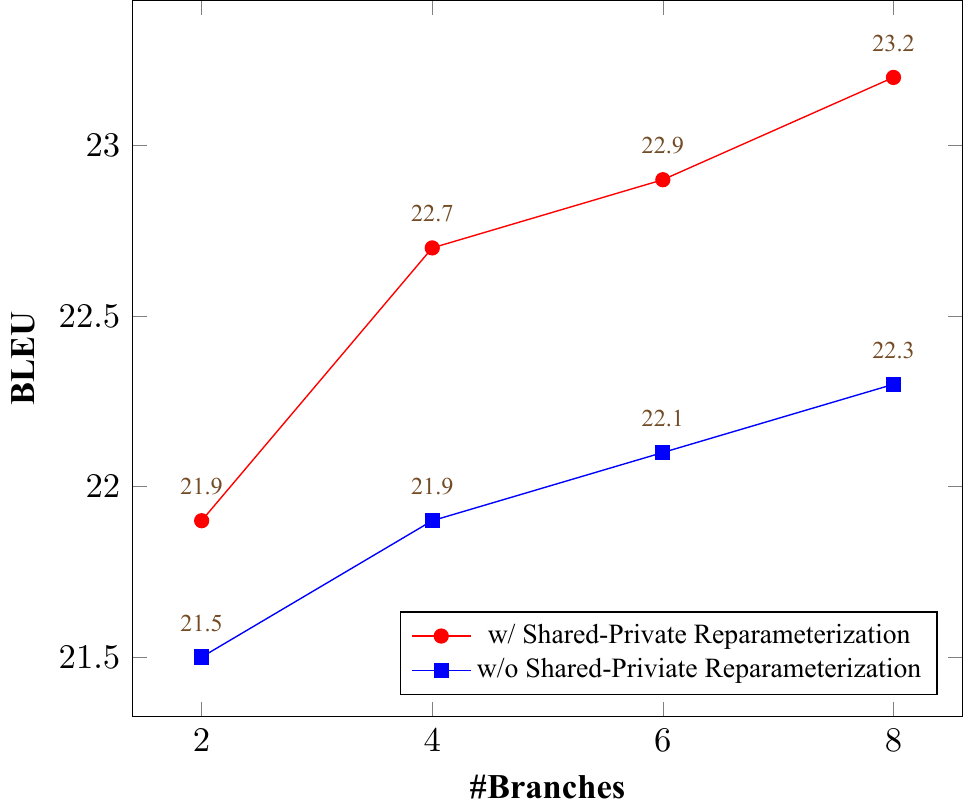}
\caption{Effect of branch number and shared-private reparameterization on the WMT14 En-De dataset.}\label{fig:abl}
\end{figure}

\begin{table*}[t]
\centering
\begin{tabular}{l rrrr rrrr}
\toprule
\multirow{2}{*}{\bf Method} & \multicolumn{4}{c}{\bf En-De} & \multicolumn{4}{c}{\bf Zh-En} \\
\cmidrule(lr){2-5} \cmidrule(lr){6-9}
& \textit{\#Params.} & \textit{\#MA.} & \textit{BLEU} & \textit{PTR} & \textit{\#Params.} & \textit{\#MA.} & \textit{BLEU} & \textit{PTR} \\
\midrule
Transformer (big)~\tiny{\cite{vaswani2017attention}} & 214M & 6.6G & 28.7 & 3.5 & 274M & 6.3G & 28.4 & 3.6 \\ \midrule
Transformer (tiny)~\tiny{\cite{vaswani2017attention}} & 7.5M & 229.0M & 23.0 & 15.2 & 11.0M & 209.7M & 21.7 & 15.0 \\
Transformer-MoE (tiny) & 15.8M & 313.2M & 24.9 & 14.1 & 19.3M & 293.9M  & 23.3 & 13.6 \\
Transformer-DMB (tiny) & 15.8M & 229.6M & 24.8 & \textbf{16.4} & 19.3M & 210.3M & 23.4 & \textbf{16.1} \\ \midrule
Transformer (small)~\tiny{\cite{vaswani2017attention}} & 20.5M & 623.2M & 26.7 & 10.7 & 27.4M & 584.6M & 26.4 & 10.9 \\
Transformer-MoE (small) & 53.7M & 956.6M & 27.1 & 8.8 & 60.6M & 918.1M & 27.2 & 9.0 \\
Transformer-DMB (small) & 53.7M & 624.3M & 27.7 & \textbf{11.1} & 60.6M & 585.7M & 26.9 & ~~\textbf{11.1} \\
\bottomrule
\end{tabular}
\caption{Experiments on WMT14 En-De and WMT20 Zh-En datasets \emph{with knowledge distillation}. ``Transformer (big)'' is used as the teacher model in knowledge distillation. ``\#Params.'' denotes the number of parameters, ``\#MA.'' denotes the number of Multi-Adds, and ``PTR'' denotes the performance-time ratio. The number of branches $N$ is set to 4 for MoE and DMB.}
\label{tab:result-kd}
\end{table*}

\begin{figure*}[!t]
\centering 
    \begin{subfigure}[b]{0.47\textwidth}
        \centering
        \includegraphics[width=\textwidth]{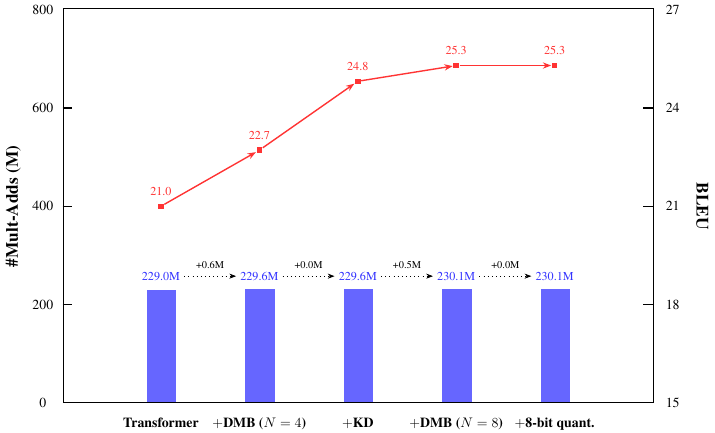}
        \caption{}
    \end{subfigure}
    \hfill
    \begin{subfigure}[b]{0.47\textwidth}
        \centering
        \includegraphics[width=\textwidth]{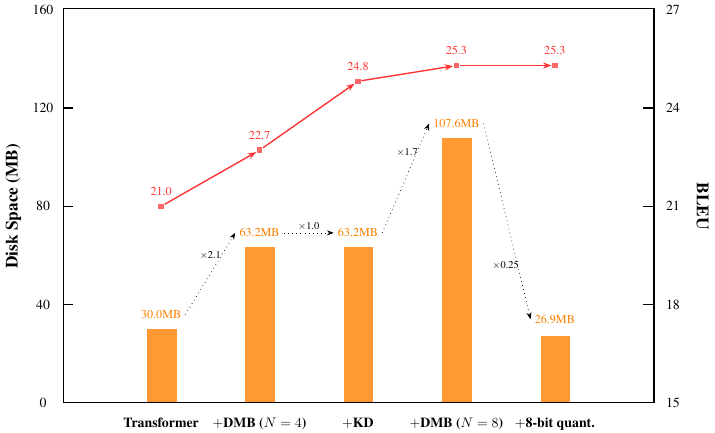}
        \caption{}
    \end{subfigure}
    \caption{(a) \#Mult-Adds vs. BLEU and (b) Disk space vs. BLEU on the WMT14 En-De translation tasks. Combined with DMB, knowledge distillation, and 8-bit quantization, we are able to achieve an improvement of 4.3 BLEU points over the Transformer model with nearly the same number of Mult-Adds and disk spaces. ``quant.'' indicates quantization.}\label{fig:compare}
\end{figure*}

We investigate the effect of the number of branches and the effectiveness of the shared-private reparameterization on the WMT14 En-De translation task. Figure \ref{fig:abl} shows the results with different settings. We have two observations: 
\begin{itemize}
\item The number of branches is a key factor that affects the performance of our Transformer-DMB models. The performance of our methods improves as the number of branches increases. This validates our assumption that the model capacity is a key bottleneck for current on-device NMT models. Our methods can significantly increase the model capacity while introducing negligible computational overhead.
\item The shared-private reparameterization is very effective. Under all settings, we can see that the model with shared-private reparameterization outperforms the corresponding counterpart. We can see that the gap between models with/without shared-private reparameterization becomes larger as $N$ increases, which proves that shared-private reparameterization can effectively address the shrinking training examples problem.
\end{itemize}

\subsection{Effect of Knowledge Distillation and Quantization}

As knowledge distillation (KD)~\cite{kim2016sequence} is a commonly used technique to improve the performance of small models, we conduct additional experiments to investigate the effectiveness of our method on the data generated with sequence-level knowledge distillation.

On the En-De translation task, our model outperforms the Transformer model over 1.8 BLEU points under tiny model settings, and 1.0 BLEU point under small model settings. The results of the Zh-En translation task are similar. Our model outperforms the Transformer model over 1.7 BLEU points under tiny model settings, and 0.5 BLEU points under small model settings. The performance of our model is also on par with the Transformer model extended with MoE. The results suggest that our method is still effective when combined with knowledge distillation.

One potential concern of the DMB approach is the increase of memory requirements. Fortunately, this problem can be alleviated with quantization techniques~\cite{chung2020extremely}.  Quantization is a very effective approach to reduce the memory consumption of NMT models, and it also improves the translation speed by utilizing integer instructions instead of floating-point operations. We show our model is orthogonal with quantization techniques. We further conduct experiments with a tiny Transformer-DMB model on the WMT14 En-De dataset with 8-bit quantization. The model has 8 branches and is trained on the data generated by knowledge distillation. The BLEU score of the quantized model is 25.3, which is the same as the model without quantization.

Figure~\ref{fig:compare} shows the BLEU, \#Mult-Adds and disk spaces of models with DMB, knowledge distillation, and quantization. Combining all techniques, we achieve 25.3 BLEU points with only 230.1M Mult-Adds. The model trained with 8-bit quantization only occupies 26.9 MB of space, which we believe is affordable for most edge devices. By using even lower bit quantization techniques~\cite{chung2020extremely}, we believe our methods can achieve a better space-time tradeoff and further improve the model performance.

\subsection{Analyses}
\subsubsection{Size of Training Data Size}
\begin{figure}[!t]{}
\centering
\includegraphics[width=0.4\textwidth]{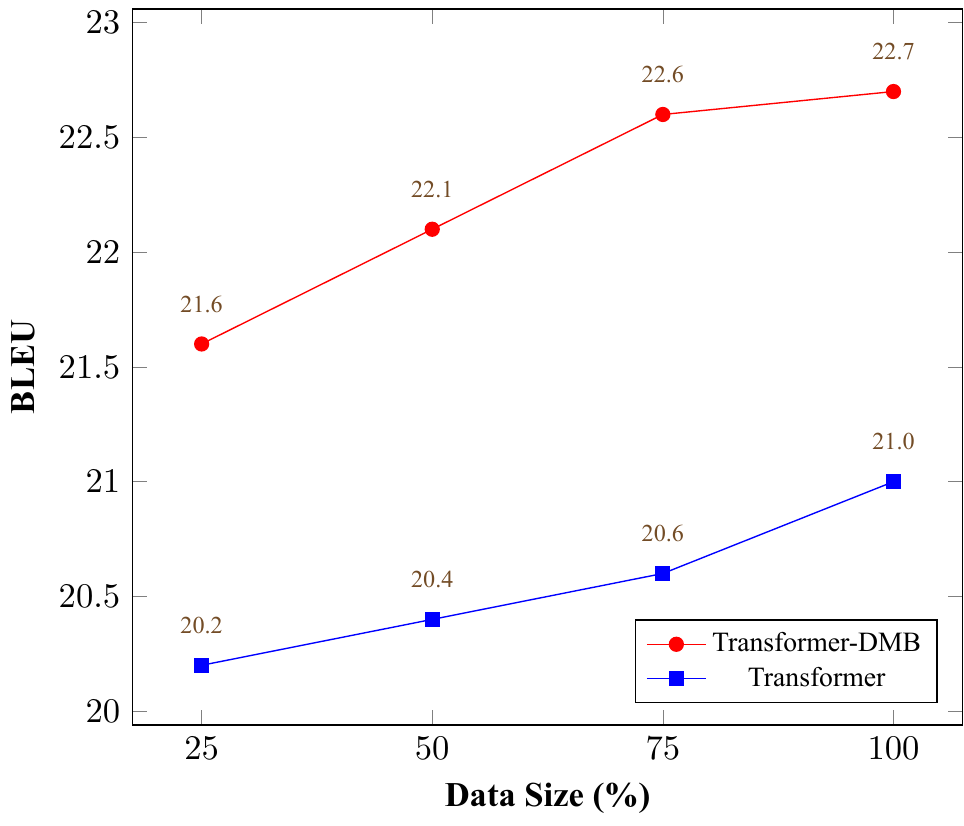}
\caption{Comparisons between Transformer-DMB models and Transformer models on the WMT14 En-De translation task with different data sizes.}\label{fig:data}
\end{figure}

It is well-known that increased capacity is helpful for neural models to absorb more training data~\cite{shazeer2017outrageously}. Therefore, we conduct experiments to study the comparisons between Transformer-DMB models (tiny) and Transformer models (tiny) at different data sizes. Figure~\ref{fig:data} shows the results. To our surprise, we find that Transformer-DMB models always outperform Transformer models regardless of the input data size. The results indicate that increasing the capacity of the model not only helps learn from more data but also is helpful for learning expressive representations.

\subsubsection{Sentence Length}

\begin{figure}[!t]{}
\centering
\includegraphics[width=0.4\textwidth]{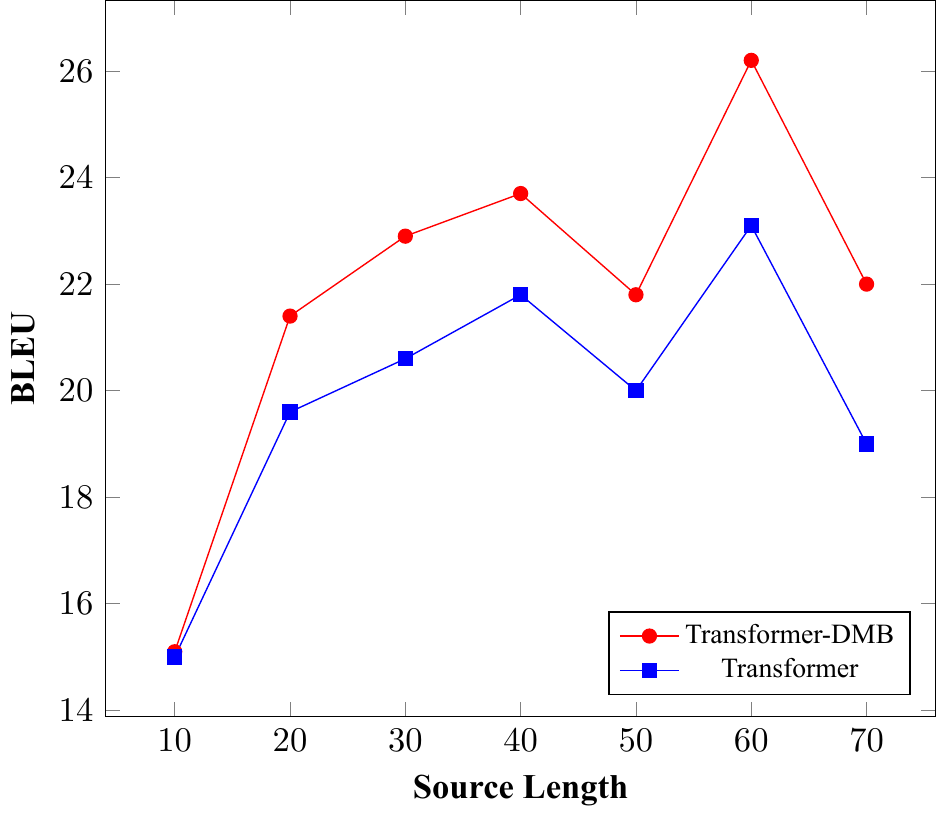}
\caption{The BLEU scores of the generated translations on the \texttt{newstest2014} with respect to the lengths of the source sentences.}\label{fig:length}
\end{figure}

As shown in Figure~\ref{fig:length}, it is clear that the Transformer-DMB model (tiny) is much better than the Transformer model (tiny) at translating long sentences. We believe this is due to the  Transformer-DMB models can use different branches to learn different semantic features of the input sentence, which is more powerful and flexible than the single-branch Transformer models.

\subsubsection{Visualization}
\begin{figure}[t]
    \centering
    \resizebox{\linewidth}{!}{\includegraphics{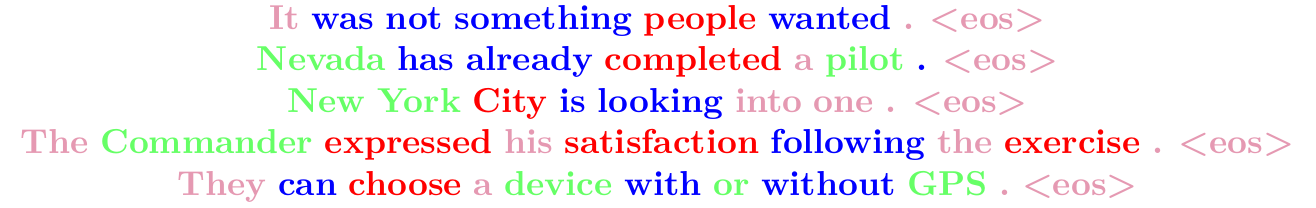}}
    \caption{Visualization of gate predictions on the En-De translation task. Different colors represent different branches.}
    \label{fig:viz}
\end{figure}

We visualize the gate predictions to find out how DMB layers process sequences. Figure~\ref{fig:viz} shows visualizations of 5 sentences. The predictions are based on the gating unit at the 4th attention sub-layer of the encoder. As we can see, different branches typically process different types of words. For example, pronouns and articles are classified into the branch highlighted in purple, named entities usually processed by the green branch. The results confirm that different branches can learn different semantic representations, thus improving the performance of NMT models.

\section{Related Work}
\paragraph{Efficient NMT} There are many works devoted to efficient NMTs, which can be roughly divided into two categories. The first line is related to model compression. Pruning~\cite{zhu2017prune,frankle2019lottery} and quantization~\cite{prato2020fully,chung2020extremely} are two widely used methods to reduce the model size. We believe our work is orthogonal with these works.
The second line focuses on reducing the computations of NMT models. \citet{wu2020lite} proposed Lite Transformer for mobile applications. They introduced a Long-Short Range Attention network to reduce the computational cost of the Transformer model. \citet{lu2020hardware} proposed the multi-head attention ResBlock and the position-wise feed-forward network ResBlock to speed up the Transformer model. \citet{so2019evolved} and \cite{wang2020hat} introduced architecture search methods to find a hardware-aware architecture. Compared with these works, our work does not need additional search time. Besides, we believe our methods are easily applicable to these new architectures.

\paragraph{Conditional Computation} \citet{bengio2013deep} first proposes the concept of conditional computation to enhance the capacity of neural networks without incurring additional computations. Various forms of conditional computation have been proposed since then~\cite{davis2013low,bengio2013estimating,eigen2013learning,cho2014exponentially,bengio2015conditional}.  \citet{shazeer2017outrageously} introduce a sparsely-gated mixture-of-experts approach. 
They build an NMT model with thousands of feed-forward neural networks between two recurrent layers. The MoE layer is similar to ours but has two major differences. First, instead of a mixture of top-k experts, we only choose one branch in a layer. Second, we introduce shared-private reparameterization in each layer during training. \citet{bapna2020controlling} propose conditional computation transformer (CCT) architecture. They add a budget loss to control the balance between quality and computation. Our work is different from theirs in several aspects. First, while they adopt an independent gating approach, we use a dynamic multi-branch approach. Second, our architecture is different from the conditional computation Transformer. Third, given a fixed sequence length, the computational costs of our models are fixed, whereas CCT depends on the complexity of the input sequence. Very recently, \citet{fedus2021switch} propose Switch Transformer architecture for scaling to trillion parameter models, which is concurrent with our work. They introduce a top-1 MoE layer which reduces the computation required in the previous approach~\cite{shazeer2017outrageously}. The differences between our DMB layers and their top-1 MoE layers are two folds: First, our gating decisions are binary-valued whereas their gating outputs are real-valued. Our method requires less computation because we do not need to multiply the gating output with the branch output. Second, we use shared-private reparameterization during training whereas they do not have a similar approach. Besides, Switch Transformer focuses on scaling up large models, while we show the effectiveness of our approach for resource-constrained applications.

\section{Conclusion}
We have proposed to use dynamic multi-branch layers to improve performance without sacrificing efficiency for on-device neural machine translation. This can be done by dynamically activating a single branch during training and inference. We also propose shared-private reparameterization for sufficient training of each branch. Experiments show that our approach achieves higher performance-time ratios than state-of-the-art approaches to on-device NMT.

\section*{Acknowledgement}
This work was supported by the National Key R\&D Program of China (No. 2018YFB1005103), National Natural Science Foundation of China (No. 62006138, No. 61925601, No. 61772302), and Huawei Noah's Ark Lab. We thank all anonymous reviewers for their valuable comments and suggestions on this work.

\bibliographystyle{named}
\bibliography{nmt}

\end{document}